\def\year{2020}\relax
\documentclass[letterpaper]{article} 
\usepackage{aaai20}  
\usepackage{times}  
\usepackage{helvet} 
\usepackage{courier}  
\usepackage[hyphens]{url}  
\usepackage{graphicx} 
\usepackage{graphicx}  
\usepackage{amsmath}
\usepackage{amssymb}
\usepackage{float}
\usepackage{mathtools}
\usepackage{algorithm}
\usepackage{algorithmic}
\usepackage{subcaption}
\usepackage{booktabs}
\usepackage{xcolor}
\usepackage{hyperref}

\title{PODNet: A Neural Network for Discovery of Plannable Options}

\author{
Ritwik Bera,\textsuperscript{\rm 1}
Vinicius G. Goecks,\textsuperscript{\rm 1}
Gregory M. Gremillion,\textsuperscript{\rm 2}
\AND
John Valasek,\textsuperscript{\rm 1} 
and Nicholas R. Waytowich\textsuperscript{\rm 2,3}\\
\textsuperscript{\rm 1}Texas A\&M University, \textsuperscript{\rm 2}Army Research Laboratory,
\textsuperscript{\rm 3}Columbia University\\
\{ritwik, vinicius.goecks, valasek\}@tamu.edu,\\
\{gregory.m.gremillion, nicholas.r.waytowich\}.civ@mail.mil
}

\begin{document}

\maketitle

\begin{abstract}
Learning from demonstration has been widely studied in machine learning but becomes challenging when the demonstrated trajectories are unstructured and follow different objectives.
This short-paper proposes PODNet, Plannable Option Discovery Network, addressing how to segment an unstructured set of demonstrated trajectories for option discovery. This enables learning from demonstration to perform multiple tasks and plan high-level trajectories based on the discovered option labels.
PODNet combines a custom categorical variational autoencoder, a recurrent option inference network, option-conditioned policy network, and option dynamics model in an end-to-end learning architecture.
Due to the concurrently trained option-conditioned policy network and option dynamics model, the proposed architecture has implications in multi-task and hierarchical learning, explainable and interpretable artificial intelligence, and applications where the agent is required to learn only from observations.

\end{abstract}

\section{Introduction}

Learning from demonstrations to perform a single task has been widely studied in the machine learning literature \cite{argall2009survey,ross2011reduction,ross2013learning,bojarski2016end,goecks2018efficiently}.
In these approaches, demonstrations are carefully curated in order to exemplify a specific task to be carried out by the learning agent.
The challenge arises when the demonstrator is performing more than one task, or multiple hierarchical sub-tasks of a complex objective, also called \emph{options}, where the same set of observations can be mapped to a different set of actions depending on the option being performed \cite{sutton1999between,stolle2002learning}.
This is a challenge for traditional behavior cloning techniques that focus on learning a single mapping between observation and actions in a single option scenario.

This paper presents Plannable Option Discovery Network (PODNet), attempting to enable agents to learn the semantic structure behind those complex demonstrated tasks by using a meta-controller operating in the option-space instead of directly operating in the action-space. 
The main hypothesis is that a meta-controller operating in the option-space can achieve much faster convergence on imitation learning and reinforcement learning benchmarks than an action-space policy network due to the significantly smaller size of the option-space.
Our contribution, PODNet, is a custom categorical variational autoencoder \cite{Jang2016} that is composed of several constituent networks that not only segment demonstrated trajectories into options, but concurrently trains an option dynamics model that can be used for downstream planning tasks and training on simulated rollouts to minimize interaction with the environment while the policy is maturing.
Unlike previous imitation-learning based approaches to option discovery, our approach does not require the agent to interact with the environment in its option discovery process as it trains offline on just behavior cloning data.
Moreover, being able to infer the option label for the current behavior executed by the learning agent, essentially, allowing the agent to broadcast the option it is currently pursuing, has implications in explainable and interpretable artificial intelligence.

\section{Related Work}


This work addresses how to segment an unstructured set of demonstrated trajectories for option discovery.
The one-shot imitation architecture developed by \citeauthor{Wang2017} \cite{Wang2017} using conditional GAIL (cGAIL) maps trajectories into a set of latent codes that capture the semantics and context of the trajectories.
This is analogous to word2vec \cite{mikolov2013efficient} in natural language processing (NLP) where words are embedded into a vector space that preserves linguistic relationships.

\begin{figure*}[!ht]
    \centering
    \includegraphics[width=0.7\textwidth]{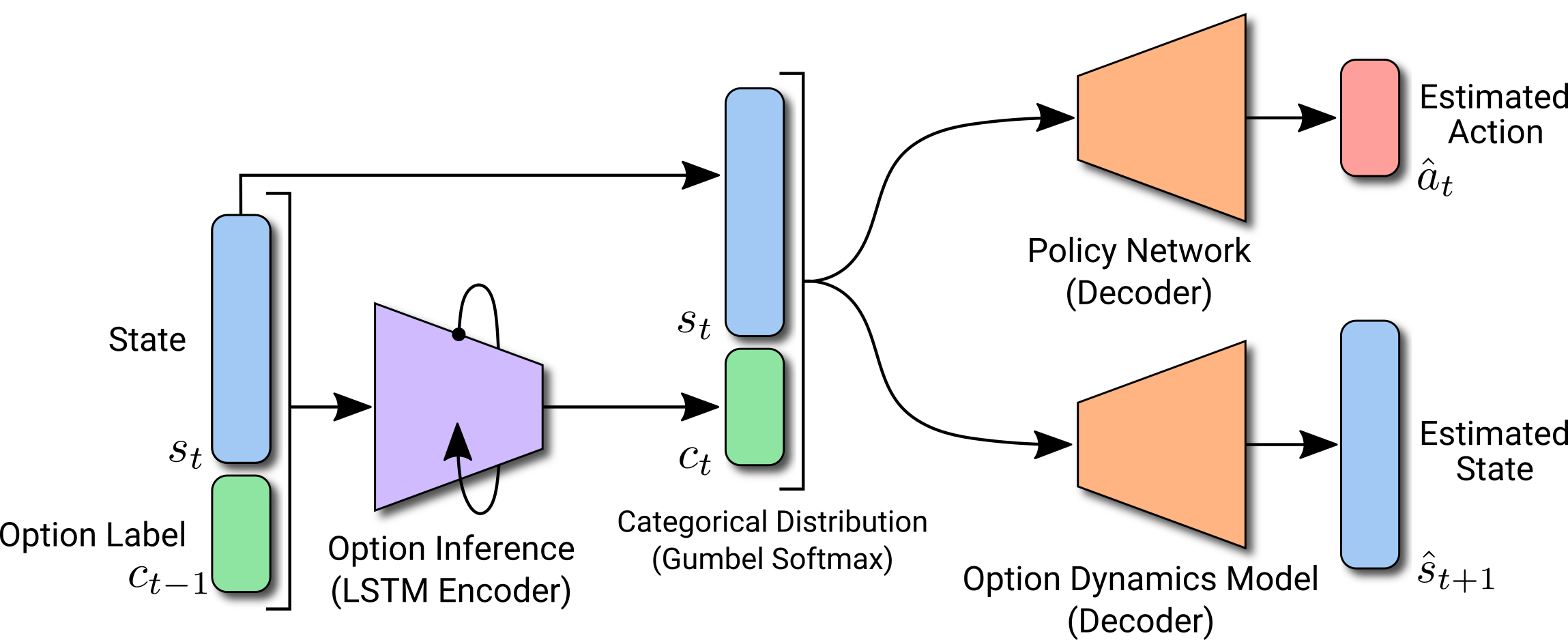}
    \caption{Proposed encoder-decoder architecture. Note that the Policy Network decoder could also be a recurrent neural network (RNN) if we wish to make the behavior label dependent on all preceding states and labels instead of just the previous state and corresponding behavior label.}
    \label{fig:vae_prop}
\end{figure*}

In InfoGAN \cite{chen2016infogan}, a generative adversarial network (GAN) maximizes the mutual information between the latent variables and the observation, learning a discriminator that confidently predict the observation labels.
InfoRL \cite{Hayat2019} and InfoGAIL \cite{Li2017} utilized the concept of mutual information maximization to map latent variables to solution trajectories (generated by RL) and expert demonstrations respectively.
Directed-InfoGAIL \cite{Sharma2018} introduced the concept of directed information. It maximized the mutual information between the trajectory observed so far and the consequent option label. This modification to the InfoGAIL architecture allowed it to segment demonstrations and reproduce option. However, it assumed a prior knowledge of the number of options to be discovered.
\emph{Diversity Is All You Need (DIAYN)} \cite{eysenbach2018diversity} recovers distinctive sub-behaviors (from random exploration) by generating random trajectories and maximising mutual information between the states and the behavior label.


\emph{Variational Autoencoding Learning of Options by Reinforcement} (VALOR)  \cite{achiam2018variational} used $\beta-$VAEs \cite{higgins2017beta} to encode labels into trajectories, thus also implicitly maximising mutual information between behavior labels and corresponding trajectories. \emph{DIAYN}'s, mutual information maximisation objective function is also implicitly solved in a $\beta-$VAE setting. 
Both VAEs and InfoGANs maximize mutual information between latent states and the input data. The difference is that VAE's have access to the true data distribution while InfoGANs also have to learn to model the true data distribution.
More recently, \emph{CompILE} \cite{kipf2019compile} employed a VAE based approach to infer not only option labels at every trajectory step but also infer option start and termination points in the given trajectory. However, once inferred to be completed, options are masked out. Thus while inferring options in the future, the agent loses track of critical options that might have happened in the past.

Most of the related works mentioned so far do not learn a dynamics model, and as a result, the discovered options cannot be used for downstream planning via model-based RL techniques. In our work, we utilize the fact that the demonstration data has state-transition information embedded within the demonstration trajectories and thus can be used to learn a dynamics model while simultaneously learning options. We also present a technique to identify the number of distinguishable options to be discovered from the demonstration data. 
\section{Plannable Option Discovery Network}


Our proposed approach, Plannable Option Discovery Network (PODNet), is a custom categorical variational autoencoder \cite{Jang2016} which consists of several constituent networks: a recurrent option inference network, an option-conditioned policy network, and an option dynamics model, as seen in Figure \ref{fig:vae_prop}. The categorical VAE allows the network to map each trajectory segment into a latent code and intrinsically perform soft k-means clustering on the inferred option labels.
The following subsections explain the constituent components of PODNet.
\subsection{Constituent Neural Networks}

\subsubsection{Recurrent option inference network}
In a complex task, the choice of an option at any time depends on both the current state, as well a history of the current and previous options that have been executed. For example, in a door opening task, an agent would decide to open a door only if it had already fetched the key earlier.
We utilize a recurrent encoder using short long-term memory (LSTM) \cite{hochreiter1997long} to ensure the current option's dependence on both the current state and the preceding options is captured. This helps overcome the problem where different options that contain similar or overlapping states are mapped to the same option label, as was observed in \emph{DIAYN} \cite{eysenbach2018diversity}. Our option inference network P is an LSTM that takes as input the current state $s_t$ as well as the previous option label $c_{t-1}$ and predicts the next option label for time step $t$. 

\subsubsection{Option-conditioned policy network}

Approaches such as \emph{InfoGAIL} \cite{Li2017}, achieve the disentanglement into latent variables by imitating the demo trajectories while having access only to the inferred latent variable and not the demonstrator actions. We achieve this goal by concurrently training a option label conditioned policy network $\pi$ that takes in the current predicted option $c_t$ as well as the current state $s_t$ and predicts the action $a_t$ that minimizes the behavior cloning loss  $\mathcal{L}_{BC}$ of the demonstration trajectories.

\begin{figure*}[!ht]
    \centering
    \includegraphics[width=0.85\textwidth]{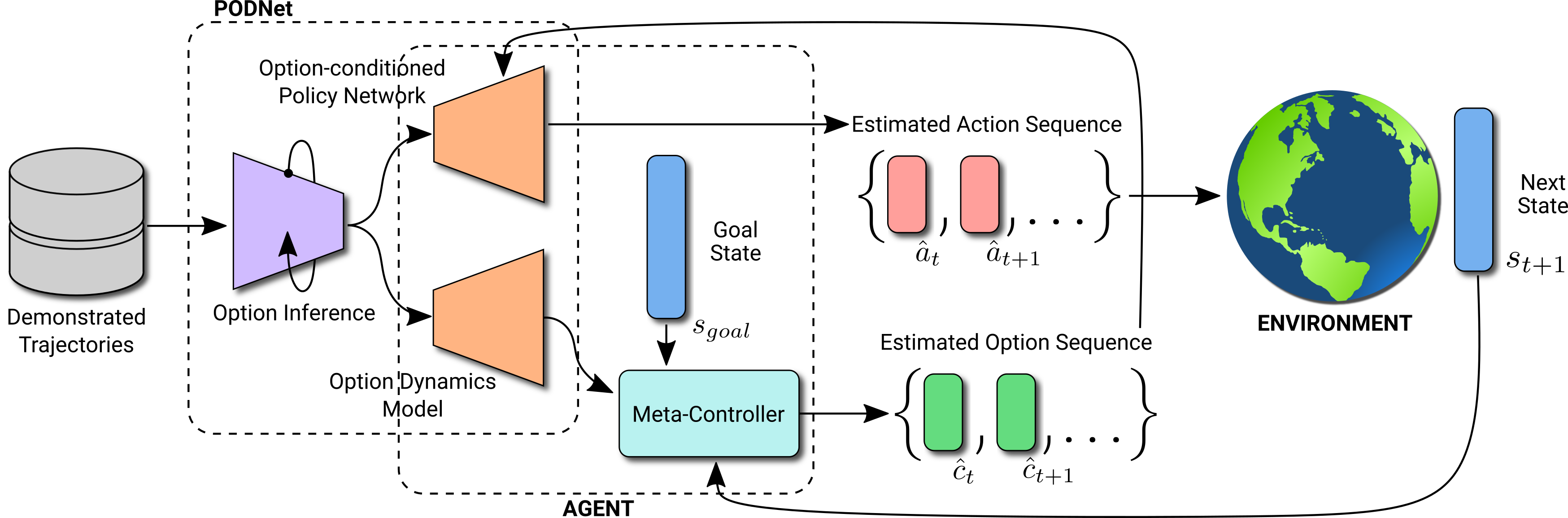}
    \caption{Complete PODNet diagram illustrating how the option dynamics model is integrated to meta-controllers to plan trajectories. Given a goal state $s_{goal}$, the meta-controller simulate trajectories using the option dynamics model and output the best estimated sequence of options to achieve the goal state. This sequence is then passed to the option-conditioned policy network, which outputs the sequence of estimated actions required to follow the planned option sequence.}
    \label{fig:vae_full}
\end{figure*}

\subsubsection{Option dynamics model}

The main novelty of PODNet is the inclusion of an option dynamics model. The options dynamics model Q takes in as input the current state $s_t$ and option label $c_t$ and predicts the next state $s_{t+1}$. In other words, the option dynamics model is an option-conditioned state-transition function, or dynamics model, that is dependent on the current option being executed instead of using the current action as traditional state-transition models would. The option dynamics model is trained simultaneously with the other policy and option inference networks by adding the \emph{option dynamics consistency loss} to the overall training objective. 
The benefit of training an option dynamics model in this way is twofold: first, it ensures the the system dynamics can be completely defined by the option label, potentially allowing for easier recovery of option labels; second, it ensures that the recovered option labels $c_t$ allow for modeling the environment dynamics in terms of the options themselves. This not only provides the ability to incorporate planning, but it allows planning to be performed at the options level instead of the action level, which will allow for more efficient planning on longer time-scales.  

\subsection{Training}

The training process occurs offline and starts by collecting a dataset $D$ consisting of unstructured demonstrated trajectories, which can be generated from any source as, for example, human experts, optimal controllers, or pre-trained reinforcement learning agents.
The overall training loss-function is given as,
\begin{align*}
    \mathcal{L}(\theta, \phi, \psi) = \mathbb{E}_{\pi_E}[\mathbb{E}_{ c_t\sim P_{\psi}(.\mid s_t, c_{t-1})} [(s_{t+1}-Q_{\phi}(s_t, c_t))^2+ \\
    (a_t-\pi_{\theta}(s_t, c_t))^2] - \beta D_{KL}{(P_{\psi}(c_t \mid s_t, c_{t-1}) \mid\mid p(c))} 
    ] 
\end{align*}

Hence,

\begin{align*}{}
\mathcal{L}(\theta, \phi, \psi) = \text{Option Dynamics Consistency Loss}(\mathcal{L}_{ODC}) +\\
\text{Behavior Cloning Loss}(\mathcal{L}_{BC}) + \text{Entropy Regularization} 
\end{align*}

\subsubsection{Ensuring smooth backpropagation}
To ensure that the gradients flow through differentiable functions only during backpropagation, $c_t$ is represented by a Gumbel-Softmax distribution, as illustrated in the literature on Categorical VAEs \cite{Jang2016}. Using argmax to select the option with highest conditional probability would lead to having a discrete operation in the neural network and prohibit backpropagation in PODNet.
Solo softmax is only used during the backward pass to allow backpropagation. For the forward pass, the softmax output is further subject to the argmax operator to obtain a one-hot encoded label vector.

\subsubsection{Entropy Regularization}
The categorical distribution arising from the encoder network is forced to have minimal KL divergence with a uniform categorical distribution. This is done to ensure that all inputs are not encoded into the same sub-behavior cluster and are meaningfully separated into separate clusters. Entropy-driven regularization encourages exploration of the label space. This exploration can be modulated by tuning the hyperparameter $\beta$.




\subsubsection{Downsampling of demonstration data}
For accurate prediction of option labels that concur with human intuition, it is important to downsample the state sequences since high-level dynamic changes have a low-frequency. Downsampling also decreases training time due to fewer samples being processed.

\subsubsection{Prediction horizon}
To ensure that the option dynamics model does not simply learn an identity projection, the dynamics model is made to predict more than one time step ahead. This prediction horizon hyperparameter could be manually tuned depending on the situation.

\subsubsection{Discovery of number of options}
The number of options can be obtained by having a held-out part of the demonstration, on which the behavior cloning loss $\mathcal{L}_{BC}$ is evaluated, similar to how validation loss is. We start with an initial number of options, K, to be discovered and increment/decrement it to move towards decreasing $\mathcal{L}_{BC}$.

\subsection{Planning Option Sequences}
Although the main motivation for PODNet is to segment unstructured trajectories, the learned option dynamics model combined with the option-conditioned policy network can be used for planning option sequences.
As shown in Figure \ref{fig:vae_full}, the option dynamics model learned with PODNets can be integrated to meta-controllers to plan trajectories. Given a goal state $s_{goal}$, the meta-controller simulate trajectories using the option dynamics model and output the best estimated sequence of options to achieve the goal state. This sequence is then passed to the option-conditioned policy network, which outputs the sequence of estimated actions required to follow the planned option sequence.


\section{Conclusion}

In this paper we presented PODNet, a neural network architecture for discovery of plannable options.
Our approach combines a custom categorical variational autoencoder, a recurrent option inference network, option-conditioned policy network, and option dynamics model for end-to-end training and segmentation of an unstructured set of demonstrated trajectories for option discovery.
PODNet's architecture implicitly utilizes prior knowledge about options being dynamically consistent (plannable and representable by a skill dynamics model), being temporally extended and definitive of the agent's actions at a particular state (as enforced by a option-conditioned policy network). This leads to discovery of plannable options that enable predictable behavior in AI agents when they adapt to newer tasks in a transfer learning setting. The proposed architecture has implications in multi-task and hierarchical learning, explainable and interpretable artificial intelligence.

\bibliographystyle{aaai}
\bibliography{refs.bib}

\begin{thebibliography}{}

\bibitem[\protect\citeauthoryear{Achiam \bgroup et al\mbox.\egroup
  }{2018}]{achiam2018variational}
Achiam, J.; Edwards, H.; Amodei, D.; and Abbeel, P.
\newblock 2018.
\newblock Variational option discovery algorithms.
\newblock {\em arXiv preprint arXiv:1807.10299}.

\bibitem[\protect\citeauthoryear{Argall \bgroup et al\mbox.\egroup
  }{2009}]{argall2009survey}
Argall, B.~D.; Chernova, S.; Veloso, M.; and Browning, B.
\newblock 2009.
\newblock A survey of robot learning from demonstration.
\newblock {\em Robotics and autonomous systems} 57(5):469--483.

\bibitem[\protect\citeauthoryear{Bojarski \bgroup et al\mbox.\egroup
  }{2016}]{bojarski2016end}
Bojarski, M.; Del~Testa, D.; Dworakowski, D.; Firner, B.; Flepp, B.; Goyal, P.;
  Jackel, L.~D.; Monfort, M.; Muller, U.; Zhang, J.; et~al.
\newblock 2016.
\newblock End to end learning for self-driving cars.
\newblock {\em arXiv preprint arXiv:1604.07316}.

\bibitem[\protect\citeauthoryear{Chen \bgroup et al\mbox.\egroup
  }{2016}]{chen2016infogan}
Chen, X.; Duan, Y.; Houthooft, R.; Schulman, J.; Sutskever, I.; and Abbeel, P.
\newblock 2016.
\newblock Infogan: Interpretable representation learning by information
  maximizing generative adversarial nets.
\newblock In {\em Advances in neural information processing systems},
  2172--2180.

\bibitem[\protect\citeauthoryear{Eysenbach \bgroup et al\mbox.\egroup
  }{2018}]{eysenbach2018diversity}
Eysenbach, B.; Gupta, A.; Ibarz, J.; and Levine, S.
\newblock 2018.
\newblock Diversity is all you need: Learning skills without a reward function.
\newblock {\em arXiv preprint arXiv:1802.06070}.

\bibitem[\protect\citeauthoryear{Goecks \bgroup et al\mbox.\egroup
  }{2018}]{goecks2018efficiently}
Goecks, V.~G.; Gremillion, G.~M.; Lawhern, V.~J.; Valasek, J.; and Waytowich,
  N.~R.
\newblock 2018.
\newblock Efficiently combining human demonstrations and interventions for safe
  training of autonomous systems in real-time.
\newblock {\em CoRR} abs/1810.11545.

\bibitem[\protect\citeauthoryear{Hayat, Singh, and
  Namboodiri}{2019}]{Hayat2019}
Hayat, A.; Singh, U.; and Namboodiri, V.~P.
\newblock 2019.
\newblock {InfoRL: Interpretable Reinforcement Learning using Information
  Maximization}.

\bibitem[\protect\citeauthoryear{Higgins \bgroup et al\mbox.\egroup
  }{2017}]{higgins2017beta}
Higgins, I.; Matthey, L.; Pal, A.; Burgess, C.; Glorot, X.; Botvinick, M.;
  Mohamed, S.; and Lerchner, A.
\newblock 2017.
\newblock beta-vae: Learning basic visual concepts with a constrained
  variational framework.
\newblock {\em ICLR} 2(5):6.

\bibitem[\protect\citeauthoryear{Hochreiter and
  Schmidhuber}{1997}]{hochreiter1997long}
Hochreiter, S., and Schmidhuber, J.
\newblock 1997.
\newblock Long short-term memory.
\newblock {\em Neural computation} 9(8):1735--1780.

\bibitem[\protect\citeauthoryear{Jang, Gu, and Poole}{2016}]{Jang2016}
Jang, E.; Gu, S.; and Poole, B.
\newblock 2016.
\newblock {Categorical Reparameterization with Gumbel-Softmax}.

\bibitem[\protect\citeauthoryear{Kipf \bgroup et al\mbox.\egroup
  }{2019}]{kipf2019compile}
Kipf, T.; Li, Y.; Dai, H.; Zambaldi, V.; Sanchez-Gonzalez, A.; Grefenstette,
  E.; Kohli, P.; and Battaglia, P.
\newblock 2019.
\newblock Compile: Compositional imitation learning and execution.
\newblock In {\em International Conference on Machine Learning},  3418--3428.

\bibitem[\protect\citeauthoryear{Li, Song, and Ermon}{2017}]{Li2017}
Li, Y.; Song, J.; and Ermon, S.
\newblock 2017.
\newblock {InfoGAIL: Interpretable Imitation Learning from Visual
  Demonstrations}.

\bibitem[\protect\citeauthoryear{Mikolov \bgroup et al\mbox.\egroup
  }{2013}]{mikolov2013efficient}
Mikolov, T.; Chen, K.; Corrado, G.; and Dean, J.
\newblock 2013.
\newblock Efficient estimation of word representations in vector space.
\newblock {\em arXiv preprint arXiv:1301.3781}.

\bibitem[\protect\citeauthoryear{Ross \bgroup et al\mbox.\egroup
  }{2013}]{ross2013learning}
Ross, S.; Melik-Barkhudarov, N.; Shankar, K.~S.; Wendel, A.; Dey, D.; Bagnell,
  J.~A.; and Hebert, M.
\newblock 2013.
\newblock Learning monocular reactive uav control in cluttered natural
  environments.
\newblock In {\em 2013 IEEE international conference on robotics and
  automation},  1765--1772.
\newblock IEEE.

\bibitem[\protect\citeauthoryear{Ross, Gordon, and
  Bagnell}{2011}]{ross2011reduction}
Ross, S.; Gordon, G.; and Bagnell, D.
\newblock 2011.
\newblock A reduction of imitation learning and structured prediction to
  no-regret online learning.
\newblock In {\em Proceedings of the fourteenth international conference on
  artificial intelligence and statistics},  627--635.

\bibitem[\protect\citeauthoryear{Sharma \bgroup et al\mbox.\egroup
  }{2018}]{Sharma2018}
Sharma, A.; Sharma, M.; Rhinehart, N.; and Kitani, K.~M.
\newblock 2018.
\newblock {Directed-Info GAIL: Learning Hierarchical Policies from Unsegmented
  Demonstrations using Directed Information}.

\bibitem[\protect\citeauthoryear{Stolle and Precup}{2002}]{stolle2002learning}
Stolle, M., and Precup, D.
\newblock 2002.
\newblock Learning options in reinforcement learning.
\newblock In {\em International Symposium on abstraction, reformulation, and
  approximation},  212--223.
\newblock Springer.

\bibitem[\protect\citeauthoryear{Sutton, Precup, and
  Singh}{1999}]{sutton1999between}
Sutton, R.~S.; Precup, D.; and Singh, S.
\newblock 1999.
\newblock Between mdps and semi-mdps: A framework for temporal abstraction in
  reinforcement learning.
\newblock {\em Artificial intelligence} 112(1-2):181--211.

\bibitem[\protect\citeauthoryear{Wang \bgroup et al\mbox.\egroup
  }{2017}]{Wang2017}
Wang, Z.; Merel, J.; Reed, S.; Wayne, G.; de~Freitas, N.; and Heess, N.
\newblock 2017.
\newblock {Robust Imitation of Diverse Behaviors}.

\end{thebibliography}

\end{document}